\documentclass[twocolumn]{article}
\usepackage{microtype}
\usepackage{enumitem}
\usepackage[dvipsnames]{xcolor}
\usepackage{multirow}
\usepackage{graphicx}
\usepackage{subcaption}
\usepackage{booktabs}
\usepackage{hyperref}
\usepackage{amsmath}
\usepackage{url}
\usepackage{float}

\usepackage{geometry}
\newcommand{\imagenet}{ImageNet }
\newcommand{\cifarten}{CIFAR-10 }
\newcommand{\cifarhundred}{CIFAR-100 }
\newcommand{\cdotfill}{%
    \leavevmode\cleaders\hb@xt@.44em{\hss$\cdot$\hss}\hfill\kern\z@
}

\title{Semantic Redundancies in Image-Classification Datasets:\\
	The 10\% You Don't Need}
\author{Vighnesh Birodkar\footnote{Work Done as Google AI Resident.} \quad Hossein Mobahi \quad Samy Bengio\\
Google Research, Mountain View, CA\\
\href{mailto:vighneshb@google.com}{vighneshb@google.com} \quad \href{mailto:hmobahi@google.com}{hmobahi@google.com} \quad \href{mailto:bengio@google.com}{bengio@google.com}
}

\date{}
\begin{document}
\maketitle

\begin{abstract}

Large datasets have been crucial to the success of deep learning models in the recent years, which keep performing better as they are trained with more labelled data. While there have been sustained efforts to make these models more data-efficient, the potential benefit of understanding the data itself, is largely untapped. Specifically, focusing on object recognition tasks, we wonder if for common benchmark datasets we can do better than random subsets of the data and find a subset that can generalize on par with the full dataset when trained on. To our knowledge, this is the  \emph{first} result that can find notable redundancies in \cifarten and \imagenet datasets (at least 10\%). Interestingly, we observe semantic correlations between required and redundant images. We hope that our findings can motivate further research into identifying additional redundancies and exploiting them for more efficient training or data-collection.

\end{abstract}

\section{Introduction}

\begin{figure*}[h]
\centering
\begin{subfigure}[b]{0.21\textwidth}
\centering
\fcolorbox{Green}{white}{\includegraphics[width=0.21\linewidth]{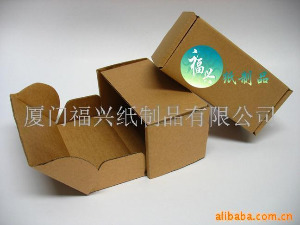}}
\includegraphics[width=0.22\linewidth]{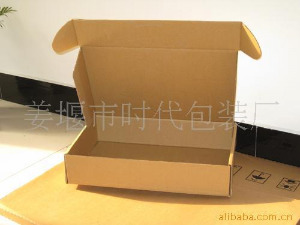}
\includegraphics[width=0.22\linewidth]{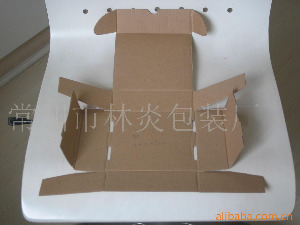}
\includegraphics[width=0.22\linewidth]{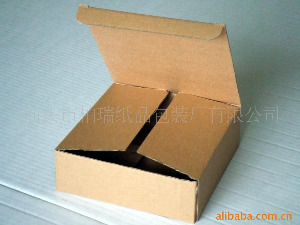}
\caption{Deformation}
\end{subfigure}
~
\begin{subfigure}[b]{0.18\textwidth}
\centering
\fcolorbox{Green}{white}{\includegraphics[width=0.4\linewidth]{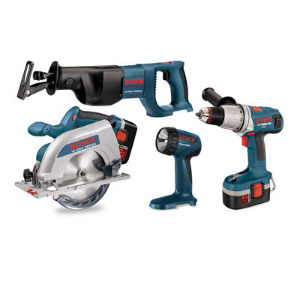}}
\includegraphics[width=0.4\linewidth]{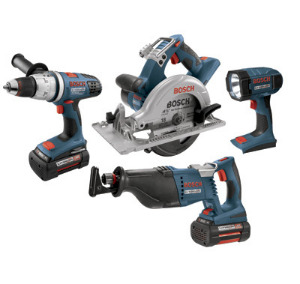}
\caption{Pose}
\end{subfigure}
~
\begin{subfigure}[b]{0.24\textwidth}
\centering
\fcolorbox{Green}{white}{\includegraphics[width=0.30\linewidth]{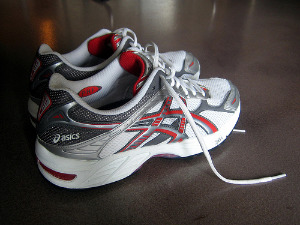}}
\includegraphics[width=0.30\linewidth]{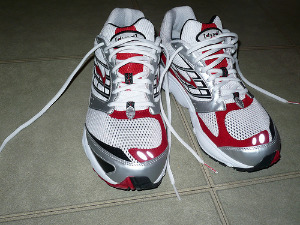}
\includegraphics[width=0.30\linewidth]{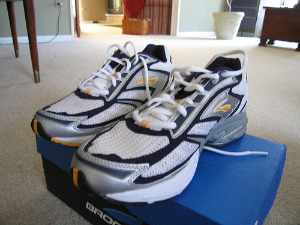}
\caption{Color, Pose}
\end{subfigure}
~
\begin{subfigure}[b]{0.24\textwidth}
\centering
\fcolorbox{Green}{white}{\includegraphics[width=0.29\linewidth]{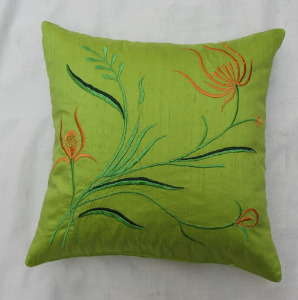}}
\includegraphics[width=0.29\linewidth]{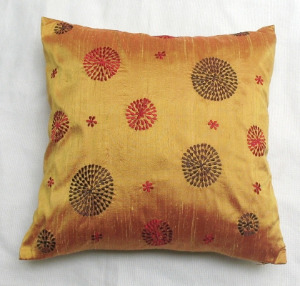}
\includegraphics[width=0.29\linewidth]{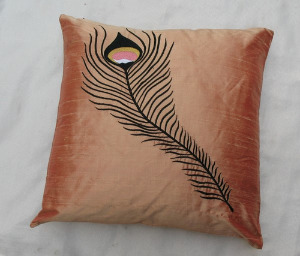}
\caption{Color, Texture}
\end{subfigure}
~

\begin{subfigure}[b]{0.21\textwidth}
\centering
\fcolorbox{Green}{white}{\includegraphics[width=0.21\linewidth]{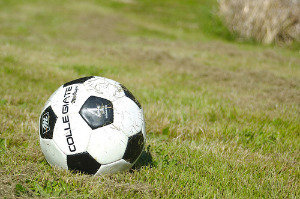}}
\includegraphics[width=0.21\linewidth]{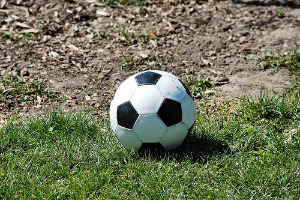}
\includegraphics[width=0.21\linewidth]{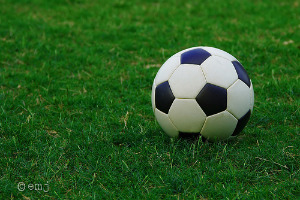}
\includegraphics[width=0.21\linewidth]{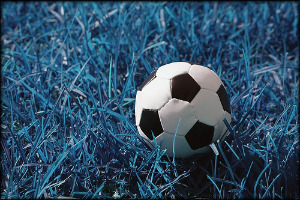}
\caption{Background}
\end{subfigure}
~
\begin{subfigure}[b]{0.18\textwidth}
\centering
\fcolorbox{Green}{white}{\includegraphics[width=0.4\linewidth]{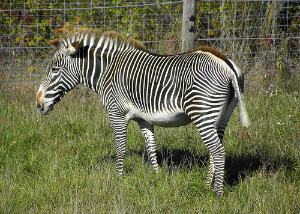}}
\includegraphics[width=0.4\linewidth]{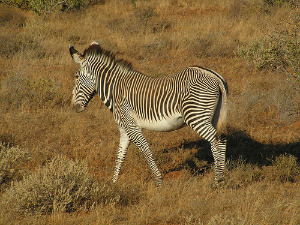}
\caption{Background}
\end{subfigure}
~
\begin{subfigure}[b]{0.24\textwidth}
\centering
\fcolorbox{Green}{white}{\includegraphics[width=0.30\linewidth]{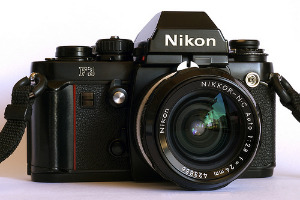}}
\includegraphics[width=0.30\linewidth]{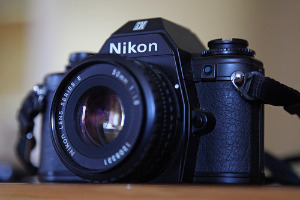}
\includegraphics[width=0.30\linewidth]{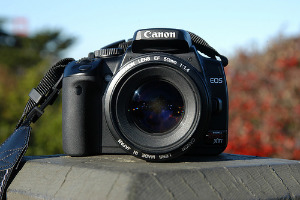}
\caption{Pose}
\end{subfigure}
~
\begin{subfigure}[b]{0.24\textwidth}
\centering
\fcolorbox{Green}{white}{\includegraphics[width=0.29\linewidth]{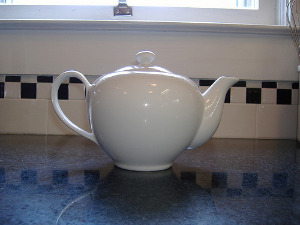}}
\includegraphics[width=0.29\linewidth]{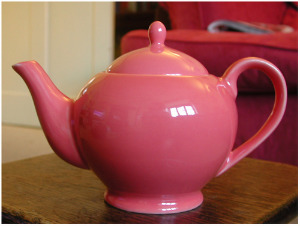}
\includegraphics[width=0.29\linewidth]{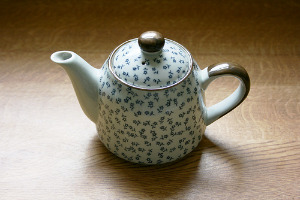}
\caption{Color, Texture, Pose}
\end{subfigure}

\caption{Examples of different redundant groups of images from the \imagenet dataset while
creating a subset 90\% of the size of the full set. In each group, we list the semantic variation
considered redundant. The images selected by semantic clustering are 
highlighted with a green box whereas the rest are discarded with
no negative impact on generalization.}
\label{fig:imagenet_dup}
\end{figure*}

Large datasets have played a central role in the recent success of deep learning. In fact, the performance of AlexNet \cite{alexnet} trained on \imagenet \cite{imagenet} in 2012 is often considered as the starting point of the current deep learning era. Undoubtedly, prominent datasets of ImageNet, CIFAR, and CIFAR-100 \cite{cifarcite} have had a crucial role in the evolution of deep learning methods since then; with even bigger datasets like OpenImages \cite{openimages} and Tencent ML-images \cite{tencent} recently emerging. These developments have led to state-of-the-art architectures such as
ResNets \cite{resnet}, DenseNets \cite{densenet}, VGG \cite{vgg}, AmoebaNets \cite{gpipe}, and regularization techniques such as Dropout \cite{dropout} and Shake-Shake \cite{shake}. However, understanding the properties of these datasets themselves has remained relatively untapped. Limited study along this direction includes \cite{focal}, which proposes a modified loss function to deal with the class imbalance inherent in object detection datasets and \cite{domain}, which studies modifications to simulated data to help models adapt to the real world, and \cite{proto} that demonstrates the existence of prototypical examples and verifies that they match human intuition.

This work studies the properties of ImageNet, \cifarten, and \cifarhundred datasets from the angle of redundancy. We find that at least 10\% of \imagenet and \cifarten can be safely removed by a technique as simple as clustering. Particularly, we identify a certain subset of \imagenet and \cifarten whose removal does not affect the test accuracy when the architecture is trained from scratch on the remaining subset. This is striking, as deep learning techniques are believed to be data hungry \cite{data1,data2}. In fact, recently the work by \cite{Malik18} specifically studying the redundancy of these datasets concludes that there is no redundancy. Our work refutes that claim by providing counter examples.

{\bf Contributions.} This work resolves some recent misconceptions about the absence of {\em notable} redundancy in major image classification datasets \cite{Malik18}. We do this by identifying a specific subset, which constitutes above 10\% of the training set, and yet its removal causes no drop in the test accuracy. To our knowledge, this is the first time such significant redundancy is shown to exist for these datasets. We emphasize that our contribution is merely to demonstrate the {\em existence} of such redundancy, but we do not claim any {\em algorithmic} contributions. However, we hope that our findings can motivate further research into identifying additional redundancies and exploiting them for more efficient training or data-collection. Our findings may also be of interest to active learning community, as it provides an upper-bound on the best performance\footnote{Suppose we learn about existence of $m$ samples in a dataset of size $n > m$ that can achieve the same test performance as a model trained with all $n$ samples. Then if our active learner cannot reach the full test performance after selecting $m$ samples, we know that there {\em might} exist a better active learning algorithm, as the ideal subset of size $m$  can achieve full test accuracy.}.

\section{Related Works}
There are approaches which try to prioritize different examples to train on as the learning process goes on such as \cite{ndf} and \cite{imp}. Although these techniques involve selecting examples to train on, they do not seek
to identify redundant subsets of the data, but rather to sample the full
dataset in a way that speeds up convergence.

An early mention of trying to reduce the training dataset size can be seen in \cite{old}. Their proposed algorithm splits the training dataset into many smaller training sets
and iteratively removes these smaller sets until the generalization performance falls below an acceptable threshold.  However, the algorithm relies on creating many small sets
out of the given training set, rendering it impractical for modern usage.

\cite{submody} pose the problem of subset selection as a constrained sub-modular
maximization problem and use it to propose an active learning algorithm. The proposed
techniques are used by \cite{diversedata} in the context of image recognition tasks. 
These drawback however, is that when used with deep-neural networks, simple
uncertainty based strategies out-perform the mentioned algorithm. 

Another example of trying to identify a smaller, more informative set
can be seen in \cite{training_value}. Using their own definition of value of
a training example, they demonstrate that prioritizing training over
examples of high training value can result in improved performance for object detection tasks. The authors suggest that their definition of training value encourages prototypicality and thus results is better learning. 

\cite{proto} attempt to directly quantify prototypicality with various metrics
and verify that all of them agree with human intuition of prototypicality
to various extents. In particular, they conclude that with
\cifarten, training on nearly-the-most
prototypical examples gives the best performance when using 10\% of the 
training data.

Most recently \cite{Malik18} attempts to find redundancies in image recognition
datasets by analyzing gradient magnitudes as a measure of importance. They prioritize
examples with high gradient magnitude according to a pre-trained classifier. Their method fails to find redundancies in
\cifarten and \imagenet
datasets.

Finally, the insights provided by our work may have implications for semi-supervised techniques assessed on notorious image datasets. Currently when evaluated on ImageNet or CIFAR datasets, a fixed-sized subset of the dataset is {\em randomly} selected according to uniform distribution, and their labels are removed \cite{ren2018metalearning, QiaoSZWY18, Tarv17, Pu16, Saj16}. This creates a training set with mix of labeled and unlabeled data to be used for assessing semi-supervised learning methods. However, creating the training set by maintain the most informative fraction of the labeled examples may provide new insights about capabilities of semi-supervised methods. 

\section{Method}
\subsection{Motivation}
In order to find redundancies, it is crucial
to analyze each sample in the context of other samples
in the dataset. Unlike previous attempts,
we seek to measure redundancy by explicitly
looking at a dissimilarity measure between samples. In
case of there being near-duplicates in the training
data, the approach of  \cite{Malik18} will
not be able to decide between them if their
resulting gradient magnitude is high, whereas
a dissimilarity measure can conclude that they
are redundant if it evaluates to a low value.
\subsection{Algorithm}
\label{sec:algo}
To find redundancies in datasets, we look at the semantic space of a pre-trained model
trained on the full dataset. In our case, the semantic representation comes
from the penultimate layer of a neural network. To find groups of points which are close by
in the semantic space we use Agglomerative Clustering
\cite{cluster}.
Agglomerative Clustering assumes that each point starts out as its own cluster initially, and at each step,
the pair of clusters which are closest
according to the dissimilarity criterion are joined together.
Given two images $I_1$ and $I_2$, whose latent representations are denoted by vectors $\boldsymbol{x}_1$ and $\boldsymbol{x}_2$. We denote the dissimilarity between $\boldsymbol{x}_1$ and $\boldsymbol{x}_2$ by $d(\boldsymbol{x}_1, \boldsymbol{x}_2)$ using the cosine angle between them as follows:
\begin{equation}
d(\boldsymbol{x}_1, \boldsymbol{x}_2) = 1 - \frac{\langle \boldsymbol{x}_1, \boldsymbol{x}_2 \rangle}
{\| \boldsymbol{x}_1 \| \, \| \boldsymbol{x}_2 \|}\;.
\label{eq:cosine}
\end{equation}
The dissimilarity between two clusters $C_1$ and $C_2$ , $D(C_1, C_2)$
is the maximum dissimilarity between any two of their constituent points:
\begin{equation}
D(C_1, C_2) = \max\limits_{\boldsymbol{x}_1 \in C_1, \boldsymbol{x}_2 \in C_2} d(\boldsymbol{x}_1, \boldsymbol{x}_2)\;.
\end{equation}

For Agglomerative Clustering, we process points belonging to each
class independently. Since the dissimilarity is a pairwise measure, processing each class separately leads to faster computations. We run the clustering algorithm until there are $k$ clusters left, where
$k$ is the size of the desired subset. 
We assume that points inside a cluster belong to the same
\emph{redundant group} of images. In each redundant group, we select
the image whose representation is closest to the cluster center and discard the rest.
Henceforth, we refer to this procedure as semantic space clustering or \emph{semantic clustering} for brevity.

\section{Experiments}
We use the ResNet \cite{resnet} architecture for all our experiments with the variant described in \cite{resnet2}.
For each dataset, we compare the performance after training on different random subsets to subsets found with semantic clustering.
Given a fixed pre-trained model,
semantic clustering subsets are deterministic and the only source
of stochasticity is due to the random network weight initialization and random mini-batch choices during
optimization by SGD.

The semantic space embedding is obtained by pre-training a network on the full
dataset. We chose the output after the last average pooling layer as our semantic space representation.
All hyperparameters are kept identical during pre-training and also when  training with different subset sizes.

As the baseline, we compare against a subset of size $k$ uniformly sampled from the full set. Each class is sampled independently to in order
to be consistent with the semantic clustering scheme. Note that random sampling scheme adds an additional source of stochasticity compared to clustering. For both either uniform sampling or cluster based subset selection, we report the mean and standard deviation of the test accuracy of the model trained from scratch using the subset.

\subsection{\cifarten \& \cifarhundred}
We train a 32-layer ResNet for the \cifarten and \cifarhundred \cite{cifarcite} datasets. The semantic representation obtained
was a $64$-dimensional vector. For both the datasets, we train
for 100,000 steps with a learning rate which is cosine annealed \cite{sgdr} from $0.1$ to $0$ with a batch size of $128$.

For optimization we use Stochastic Gradient Descent with a momentum of coefficient of $0.9$. We regularize our weights
by penalizing their $\ell_2$ norm with a factor of $0.0001$.
We found that to prevent
weights from diverging when training with subsets of all sizes, warming up the learning rate was necessary. We use linear
learning rate warm-up for $2500$ steps from $0$. We verified that warming up the learning rate performs slightly better than using no warm-up
when using the full dataset.

In all these experiments, we report average test accuracy across 10 trials.

\subsubsection{\cifarten}

\begin{figure}
\includegraphics[width=1.1\linewidth]{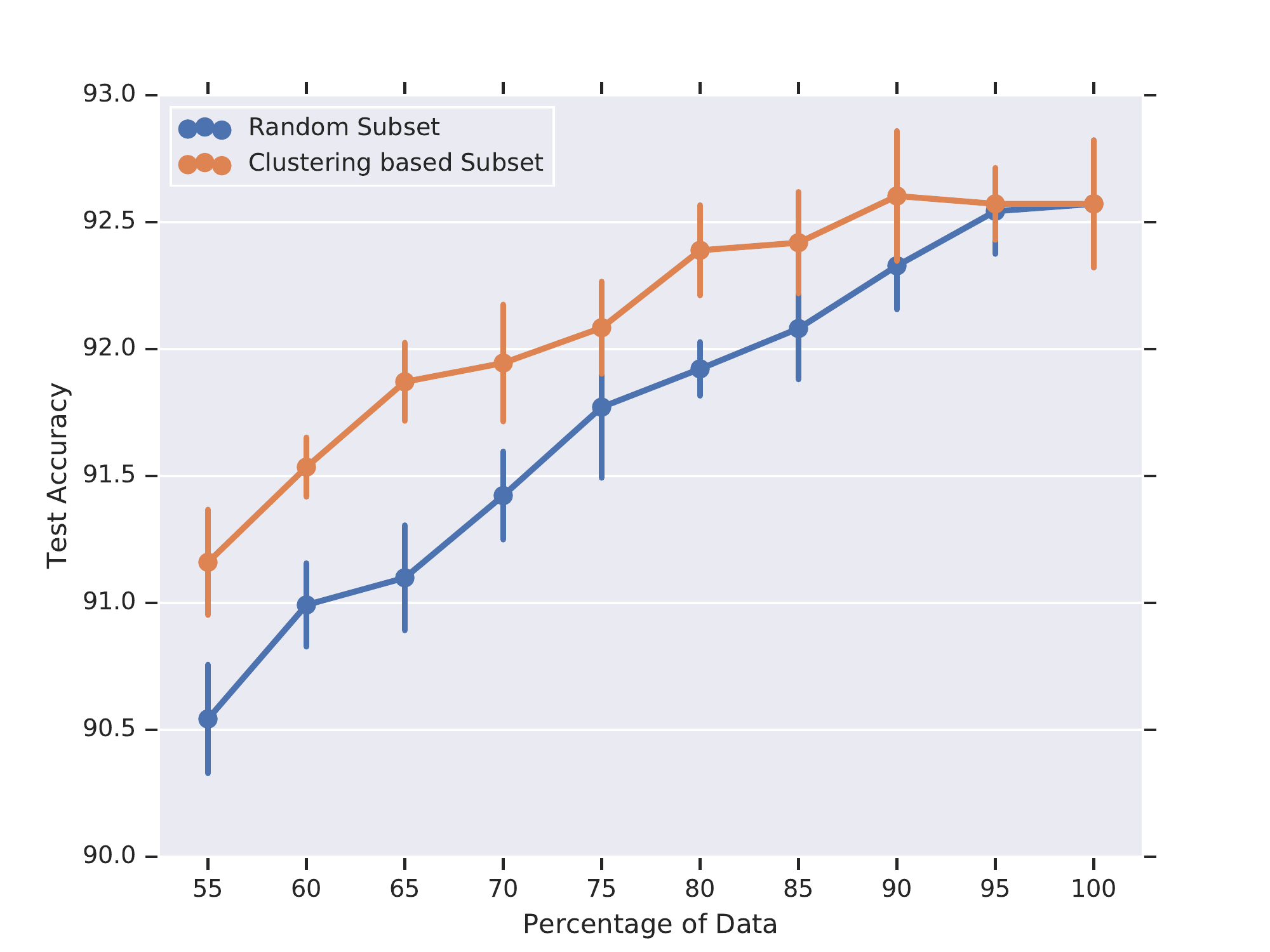}
\caption{Performance of subsets of varying size on the \cifarten dataset. Each point is an average across $10$ trials
and the vertical bars denote standard deviation.
We see no drop in test accuracy
until 10\% of the data considered redundant by semantic clustering is removed.} 
\label{fig:subset_cifar10}
\end{figure}

We see in the case of the \cifarten dataset in Figure \ref{fig:subset_cifar10} that the same test accuracy can be
achieved even after 10\% of the training is discarded using semantic clustering. In contrast, training
on random subsets of smaller sizes, results in a monotonic drop in performance. Therefore, while we show that at least 10\% of the data in the \cifarten dataset is redundant, this redundancy cannot be observed by uniform sampling.

\begin{figure}[h]
\centering

\begin{subfigure}[b]{0.28\textwidth}
\centering
\fcolorbox{Green}{White}{\includegraphics[width=0.28\linewidth]{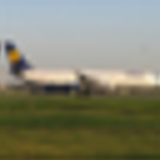}}
\includegraphics[width=0.28\linewidth]{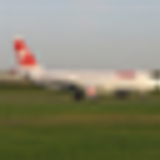} 
\includegraphics[width=0.28\linewidth]{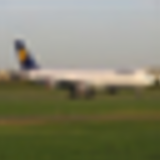} 
\caption{Redundant Airplanes}
\label{fig:red_plane1}
\end{subfigure}
~
\begin{subfigure}[b]{0.18\textwidth}
\centering
\fcolorbox{Green}{White}{\includegraphics[width=0.4\linewidth]{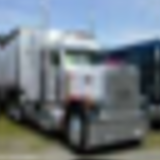}}
\includegraphics[width=0.4\linewidth]{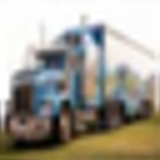} 
\caption{Redundant Trucks}
\label{fig:red_truck1}
\end{subfigure}

\begin{subfigure}[b]{0.28\textwidth}
\centering
\fcolorbox{Green}{White}{\includegraphics[width=0.28\linewidth]{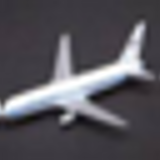}}
\includegraphics[width=0.28\linewidth]{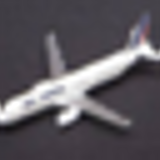} 
\includegraphics[width=0.28\linewidth]{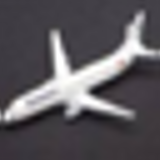} 
\caption{Redundant Airplanes}
\label{fig:red_plane2}
\end{subfigure}
~
\begin{subfigure}[b]{0.18\textwidth}
\centering
\fcolorbox{Green}{White}{\includegraphics[width=0.4\linewidth]{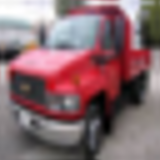} }
\includegraphics[width=0.4\linewidth]{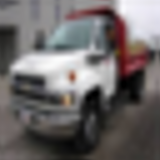} 
\caption{Redundant Trucks}
\label{fig:red_truck2}
\end{subfigure}
\caption{Examples of redundant images in the \cifarten dataset when creating
a subset of 90\% size of the original set. The figure illustrates
similarity between images of each redundant group and variation across different
redundant groups.
\ref{fig:red_plane1} and \ref{fig:red_plane2} are two different redundant
groups of the class Airplane.
\ref{fig:red_truck1} and \ref{fig:red_truck2} are two different redundant groups from
class Truck. In each group, only the images marked with green boxes are  kept and the rest, discarded.
The discarded images did not lower test accuracy.}
\label{fig:cifar_redundant}
\end{figure}

Figure \ref{fig:cifar_redundant} shows examples of images considered redundant with
semantic clustering while choosing a subset of 90\% size of the full dataset.
Each set denotes images the were placed in into the same (redundant) group by semantic clustering. Images in green boxes were retained while the rest were discarded.

\begin{figure}
\includegraphics[width=1.1\linewidth]{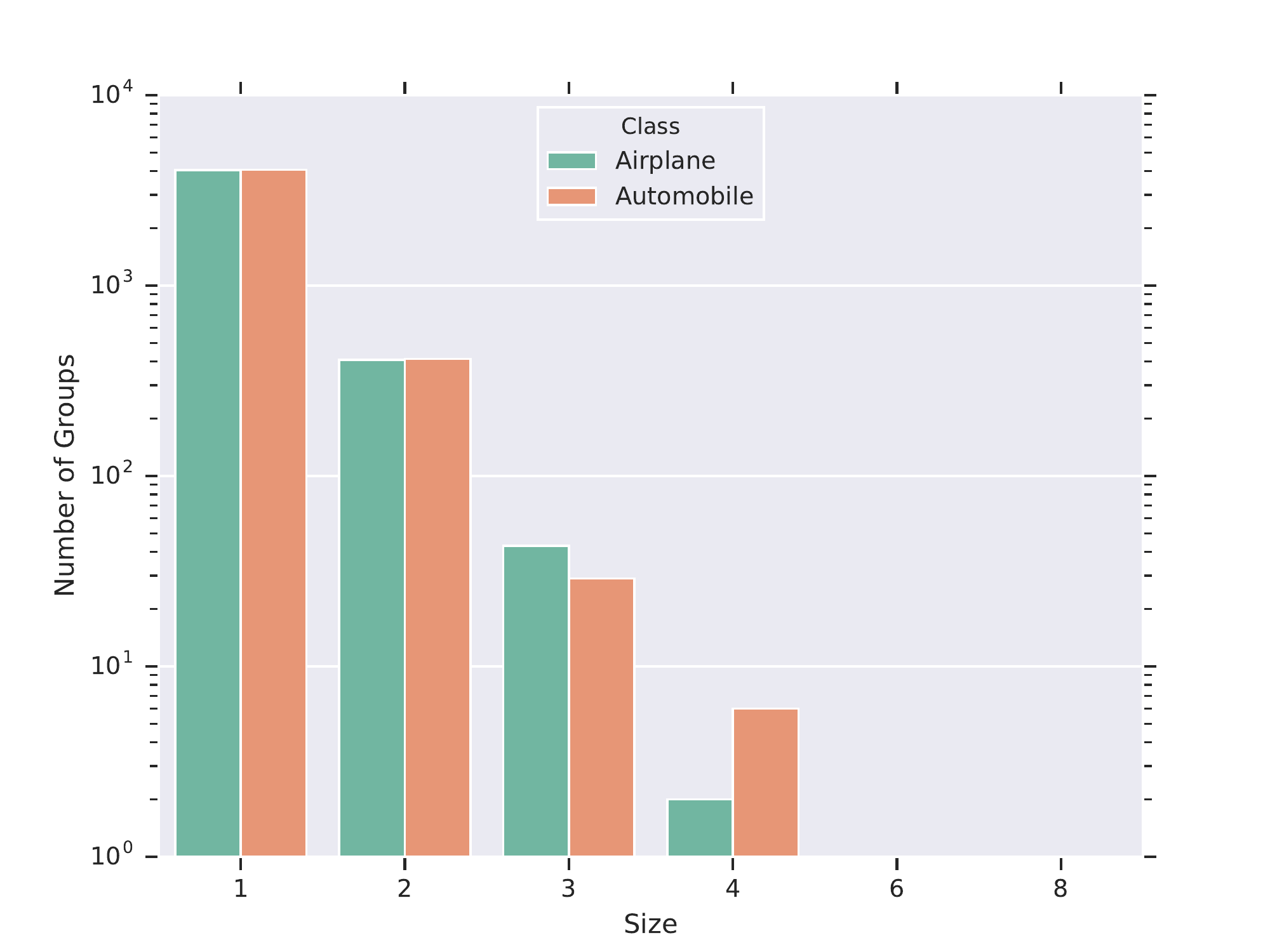}
\caption{Number of redundant groups of various sizes in the \cifarten dataset when finding a 90\% subset for two classes. Note that the y-axis
is logarithmic. }
\label{fig:sizes_cifar10}
\end{figure}

Figure \ref{fig:sizes_cifar10} shows the number of redundant
groups of different sizes for two classes in the  \cifarten dataset when seeking a 90\% subset. Since a majority
of points are retained, most clusters end up containing one element upon termination. Redundant points arise from clustering with two or more elements in them.

\subsubsection{\cifarhundred}

\begin{figure}
\includegraphics[width=1.1\linewidth]{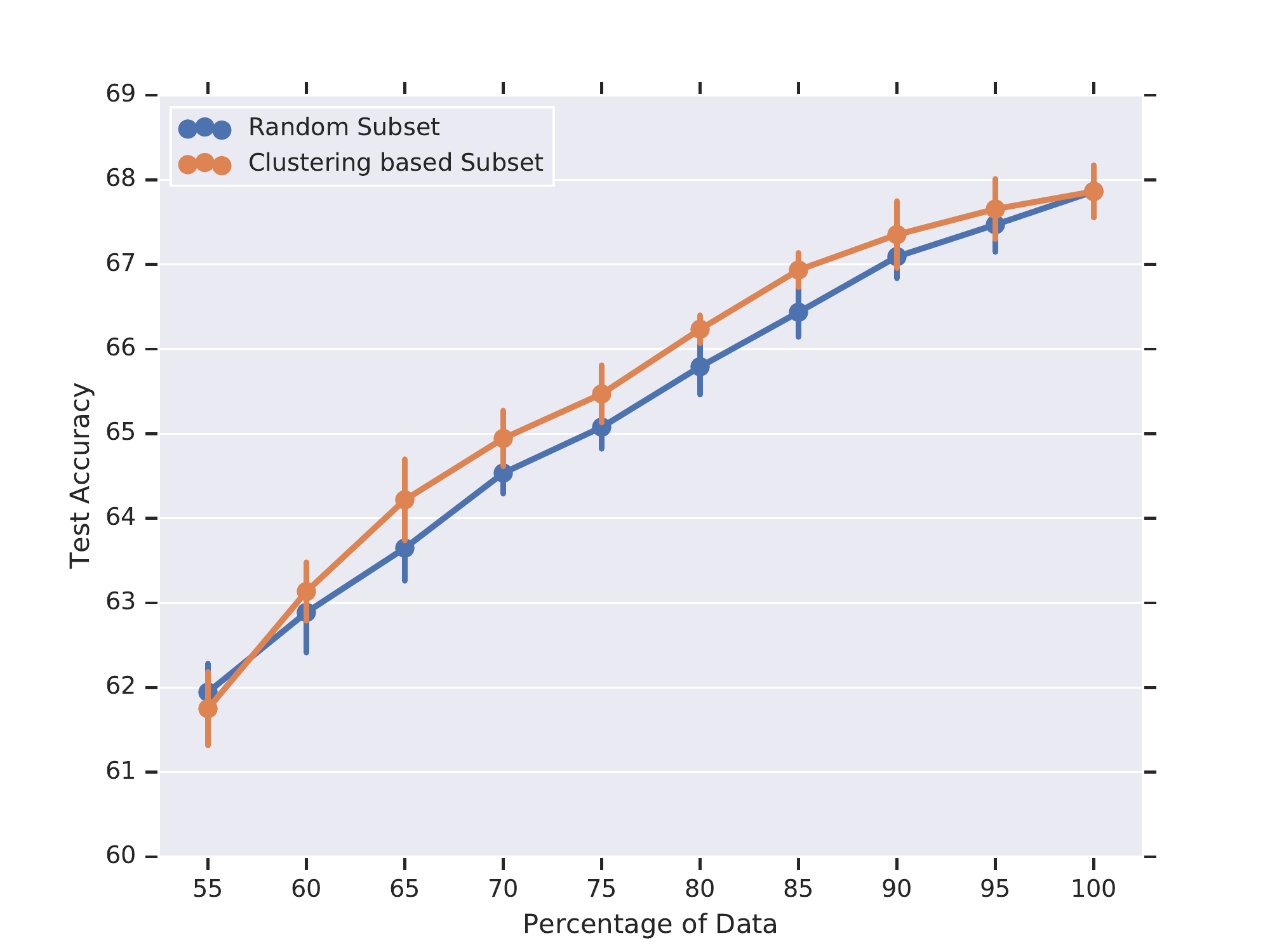}
\caption{Performance of subsets of varying size on the \cifarhundred dataset. Each point is an average over $10$ trials and the vertical
bars denote standard deviation.} 
\label{fig:subset_cifar100}
\end{figure}

In the case of the \cifarhundred dataset, our proposed scheme fails to find redundancies, as is shown in Figure \ref{fig:subset_cifar100}, while it does slightly better than random subsets. Both proposed and random methods show a monotonic decrease in test accuracy with decreasing subset size. 

\begin{figure}
\centering

\begin{subfigure}[b]{0.22\textwidth}
\centering
\captionsetup{justification=centering}

\fcolorbox{Green}{White}{\includegraphics[width=0.37\linewidth]{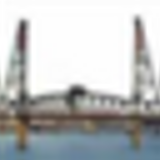}}
\includegraphics[width=0.37\linewidth]{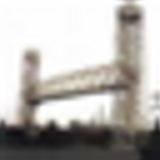} 
\caption{Redundant Bridges}
\end{subfigure}
~
\begin{subfigure}[b]{0.22\textwidth}
\centering
\fcolorbox{Green}{White}{\includegraphics[width=0.37\linewidth]{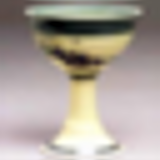}}
\includegraphics[width=0.37\linewidth]{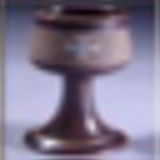} 
\caption{Redundant Cups}
\end{subfigure}

\begin{subfigure}[b]{0.22\textwidth}
\centering
\fcolorbox{Green}{White}{\includegraphics[width=0.37\linewidth]{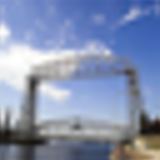}}
\includegraphics[width=0.37\linewidth]{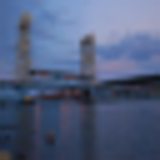} 
\caption{Redundant Bridges}
\label{fig:bridge2}
\end{subfigure}
~
\begin{subfigure}[b]{0.22\textwidth}
\centering
\fcolorbox{Green}{White}{\includegraphics[width=0.37\linewidth]{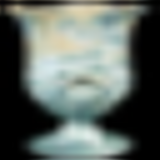}}
\includegraphics[width=0.37\linewidth]{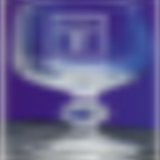} 
\caption{Redundant Cups}
\end{subfigure}

\caption{Example of variation between images in the same redundant group compared to
variation across different redundant groups in the \cifarhundred dataset. Each column
contains a specific class of images. In contrast to Figure \ref{fig:cifar_redundant}, the
images within each redundant group show much more variations. The groups were found
when retaining a 90\% subset, and retraining only the selected images (in green boxes)
and discarding the rest had a negative impact on test performance.}
\label{fig:cifar100_redundant}

\end{figure}

Figure \ref{fig:cifar100_redundant} looks at redundant groups found with semantic clustering to
retain 90\% of the dataset. As compared to Figure \ref{fig:cifar_redundant}, the images
within a group show much more semantic variation. Redundant groups in Figure \ref{fig:cifar_redundant}
are slight variations of the same object, where as
in Figure \ref{fig:cifar100_redundant}, redundant
groups do not contain the same object. We note
that in this case the model is not able to
be invariant to these semantic changes.

\begin{figure}
\includegraphics[width=1.1\linewidth]{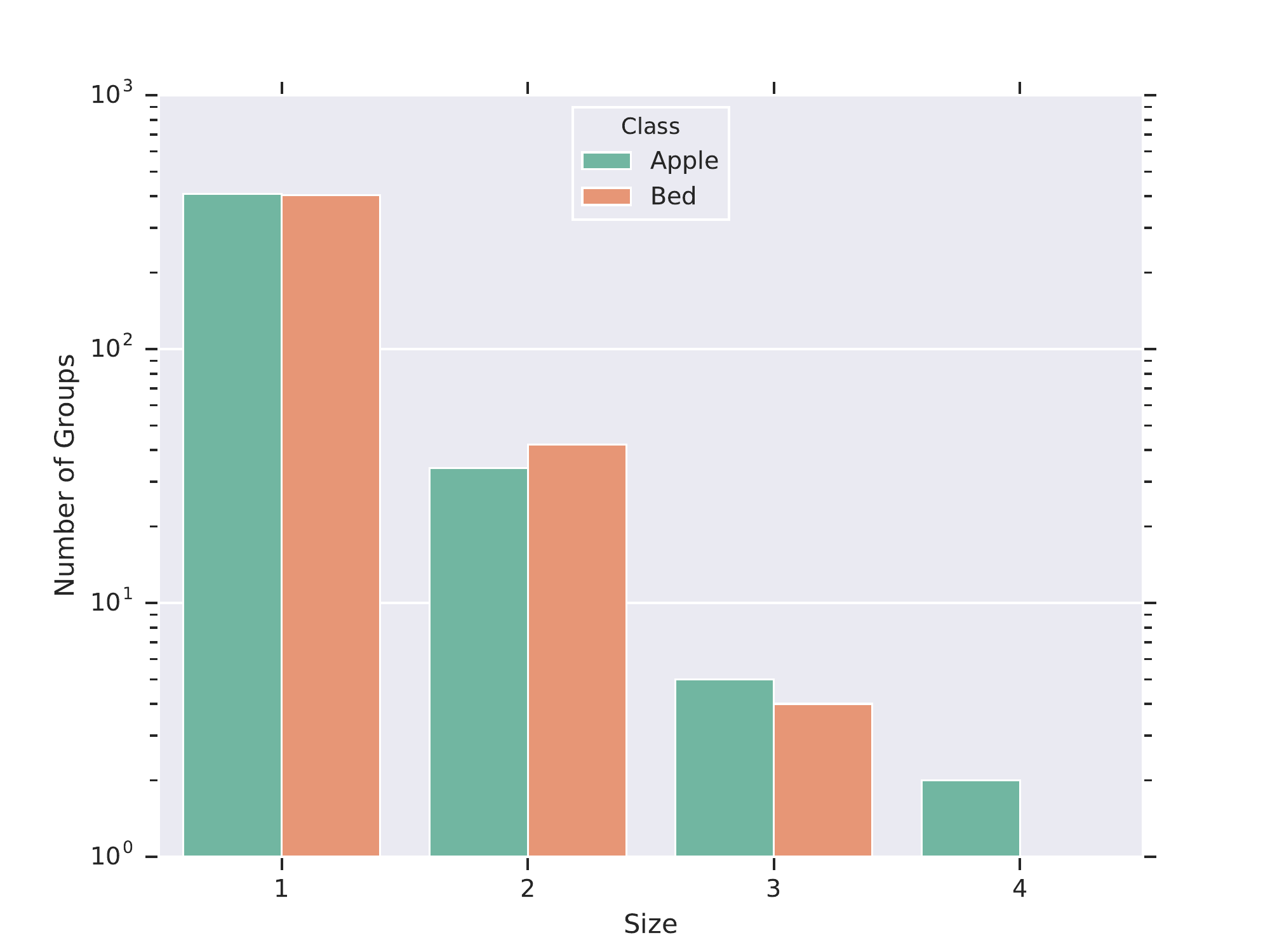}
\caption{Number of redundant groups of various sizes in the \cifarhundred dataset when finding a 90\% subset for two classes. Note that the y-axis
is logarithmic. }
\label{fig:sizes_cifar100}
\end{figure}

Similar to Figure \ref{fig:sizes_cifar10}, we 
plot the number of redundant groups of each
size for two classes in \cifarhundred in  Figure 
\ref{fig:sizes_cifar100}. 

To quantify the semantic variation of \cifarhundred
in relation to \cifarten, we select 
redundant groups of size two or more, and measure
the average dissimilarity(from Equation \ref{eq:cosine}) to the retained
sample. We report the average over groups in 3 different classes
as well as the entire datasets in Table \ref{tab:avg_dist}. It is clear that the higher
semantic variation in the redundant groups of \cifarhundred seen in Figure \ref{fig:cifar100_redundant} translates to an higher average dissimilarity in 
Table \ref{tab:avg_dist}.


\begin{table}
\centering
\begin{tabular}{llr}
\toprule
Dataset & Class & Average Dissimilarity \\
\midrule
\multirow{4}{*}{\cifarten}& Airplane & $1.73 \times 10^{-3}$ \\ 
& Automobile & $1.65 \times 10^{-3}$ \\ 
& Bird & $2.22 \times 10^{-3}$ \\ 
& {\bf All (mean)} & $1.84 \times 10^{-3}$ \\
\midrule
\multirow{4}{*}{\cifarhundred} & Apple & $6.61 \times 10^{-3}$ \\
& Bed & $14.16 \times 10^{-3}$ \\ 
& Bowl & $20.02 \times 10^{-3}$ \\ 
& {\bf All (mean)} & $13.90 \times 10^{-3}$ \\ 
\end{tabular}
\caption{Average dissimilarity to the retained sample
across redundant groups (clusters) of size greater
than 1. We report the class-wise mean for 3 classes
as well as the average over the entire dataset. 
All clusters were created to find a subset of 90\%
the size of the full set. We can observe that
the average dissimilarity is about an order of magnitude
higher for the \cifarhundred dataset, indicating
that there is more variation in the redundant groups.
}
\label{tab:avg_dist}
\end{table}

\subsection{Choice of semantic representation.}
\begin{figure}
\includegraphics[width=1.1\linewidth]{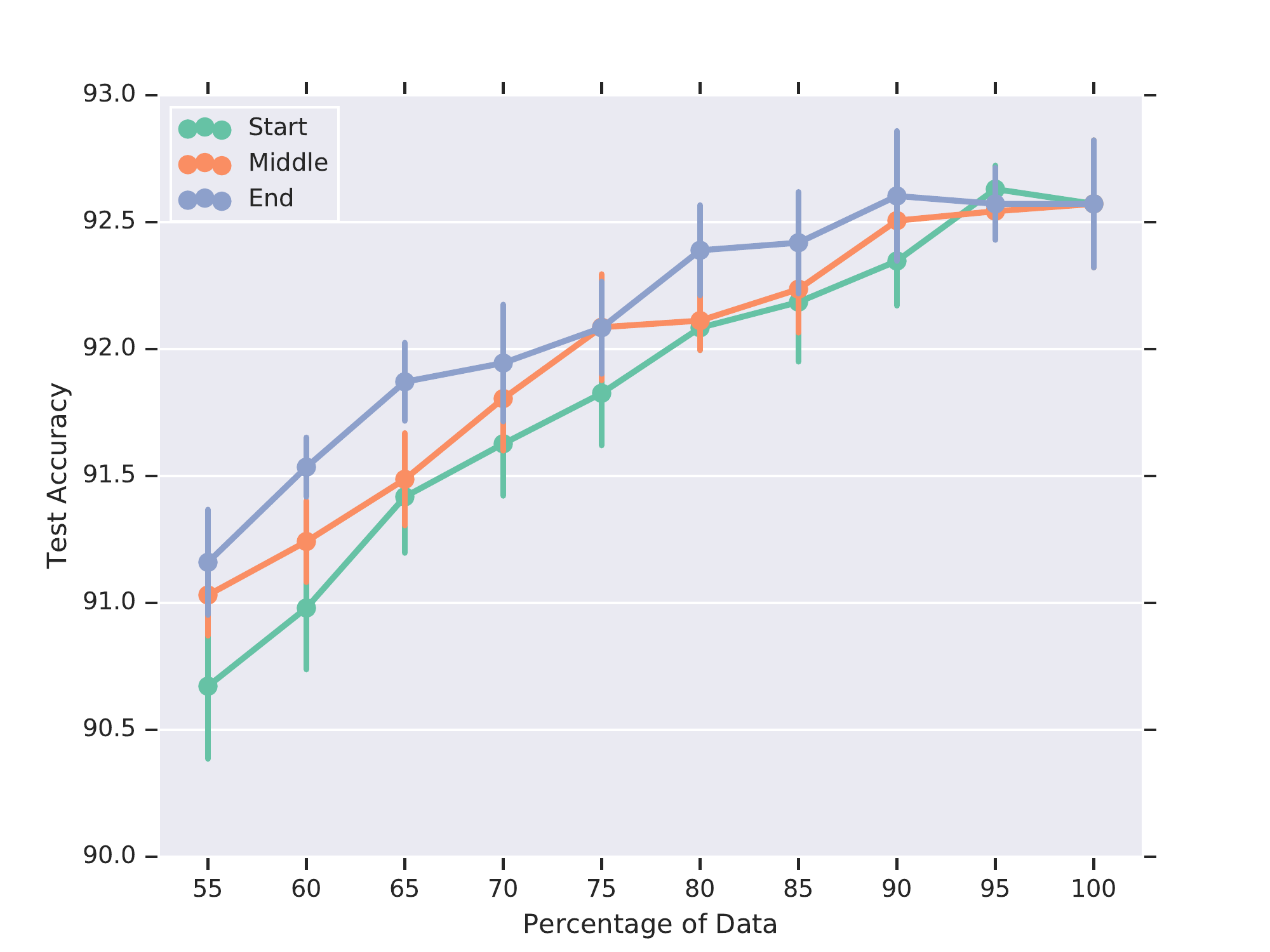}
\caption{Effectiveness of latent representations from 3 stages in a ResNet at finding redundant subsets.} 
\label{fig:layerwise_cifar10}
\end{figure}

To determine the best choice of semantic representation from a pre-trained model, we run experiments after
selecting the semantic representation from 3 different layers in the network. Figure \ref{fig:layerwise_cifar10} shows
the results. Here ``Start'' denotes the semantic representation after the first Convolution layer is a ResNet, ``Middle``
denotes the representation after the second residual block, and ``End'' denotes the output of the last average pooling
layer. We see that the ``End'' layer's semantic representation is able to find the largest redundancy.

\subsection{\imagenet}
We train a 101-layer ResNet with the \imagenet dataset. It gave us
a semantic representation of $2048$ dimensions. We use a batch size of $1024$ during training and train for $120,000$ steps with a learning
rate cosine annealed from $0.4$ to $0$. Using the strategy from \cite{largebatch}, we linearly warm up our learning rate from $0$ for 5000 steps
to be able to train with large batches. We regularize our weights with $\ell_2$ penalty with a factor of $0.0001$.

For optimization, we use
Stochastic Gradient Descent with a Momentum coefficient of $0.9$ while using the Nesterov momentum update.
Since the test set is not publicly available we report the average
validation accuracy, measured over $5$ trials.

\begin{figure}
\includegraphics[width=1.1\linewidth]{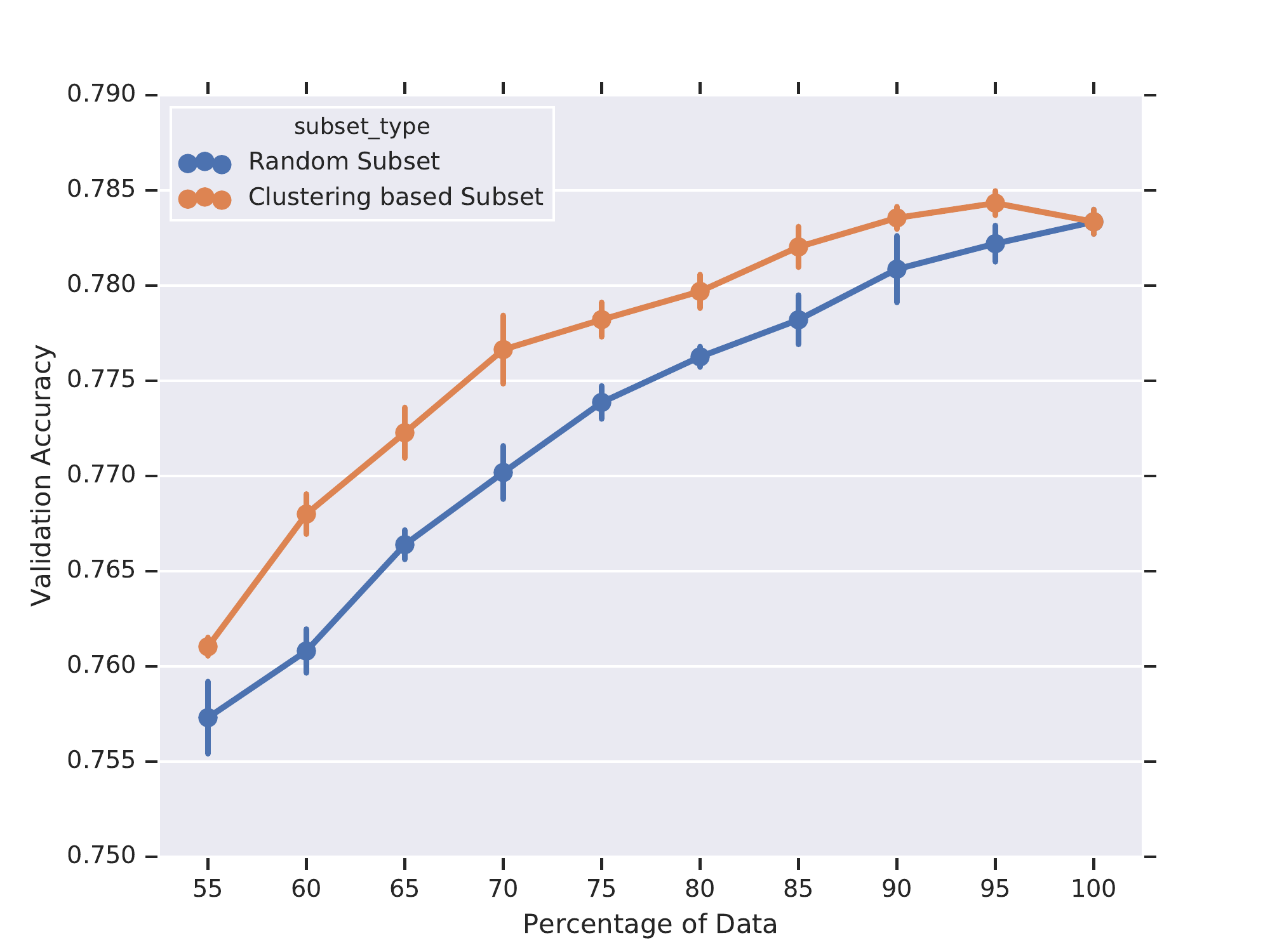}
\caption{Validation accuracy after training with subsets of various sizes of \imagenet. We plot the average over $5$ trials with
the vertical bars denoting standard deviation. There is no drop in validation accuracy when 10\% of the training data considered
redundant by semantic clustering is removed.}
\label{fig:subset_imagenet}
\end{figure}

\begin{figure}
\includegraphics[width=1.1\linewidth]{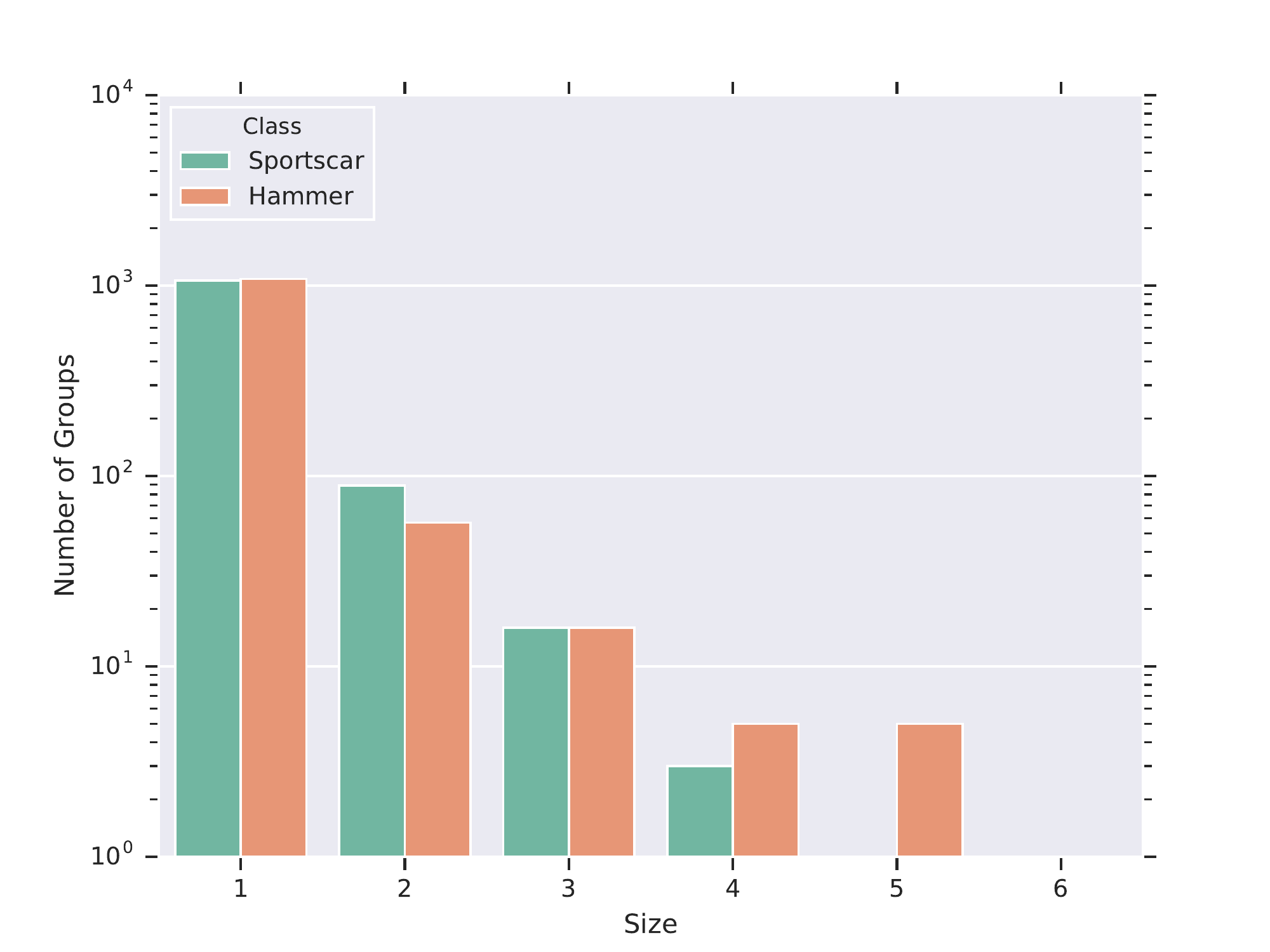}
\caption{Sizes of redundant groups for the Hammer and Sports-car classes in the \imagenet dataset when finding a 90\% subset. Note that the y-axis
is logarithmic. }
\label{fig:sizes_imagenet}
\end{figure}

\begin{figure*}[h]

\begin{subfigure}[b]{0.20\textwidth}
\centering
\fcolorbox{Green}{White}{\includegraphics[width=0.4\linewidth]{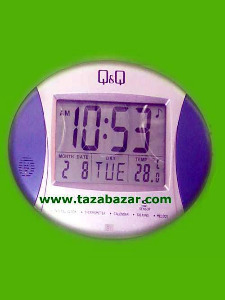}} 
\includegraphics[width=0.4\linewidth]{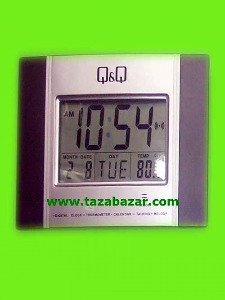}
\caption{Redundant Clocks}
\label{fig:red_clock1}
\end{subfigure}
~
\begin{subfigure}[b]{0.22\textwidth}
\centering
\fcolorbox{Green}{White}{\includegraphics[width=0.3\linewidth]{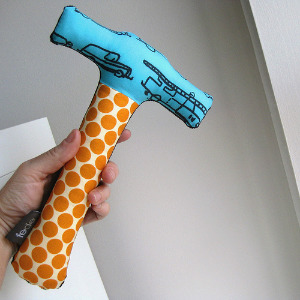}} 
\includegraphics[width=0.3\linewidth]{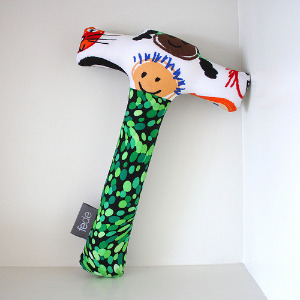} 
\includegraphics[width=0.3\linewidth]{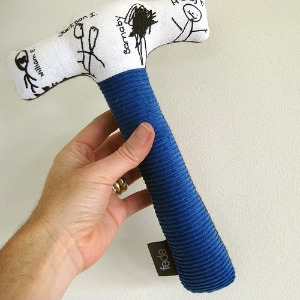} 
\caption{Redundant Hammers}
\label{fig:red_hammer1}
\end{subfigure}
~
\begin{subfigure}[b]{0.3\textwidth}
\centering
\fcolorbox{Green}{White}{\includegraphics[width=0.2\linewidth]{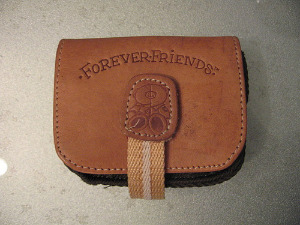}} 
\includegraphics[width=0.2\linewidth]{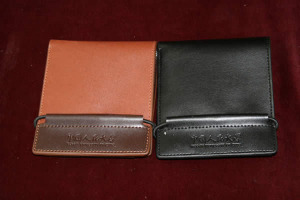} 
\includegraphics[width=0.2\linewidth]{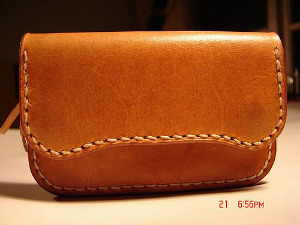} 
\includegraphics[width=0.2\linewidth]{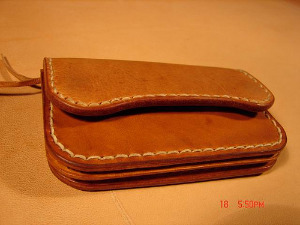}
\caption{Redundant Purses}
\label{fig:red_purses1}
\end{subfigure}
~
\begin{subfigure}[b]{0.22\textwidth}
\centering
\fcolorbox{Green}{White}{\includegraphics[width=0.3
\linewidth]{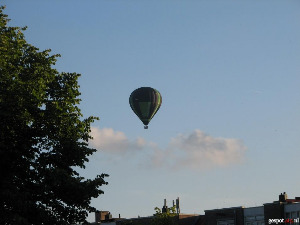}} 
\includegraphics[width=0.3\linewidth]{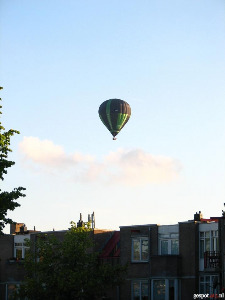}
\includegraphics[width=0.3\linewidth]{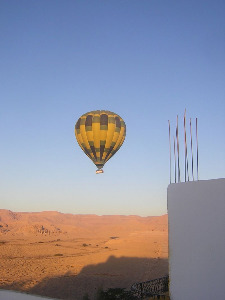}
\caption{Redundant Balloons}
\label{fig:red_balloon1}
\end{subfigure}

\begin{subfigure}[b]{0.20\textwidth}
\centering
\fcolorbox{Green}{White}{\includegraphics[width=0.4\linewidth]{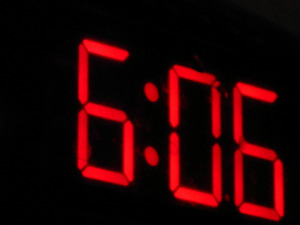}}
\includegraphics[width=0.4\linewidth]{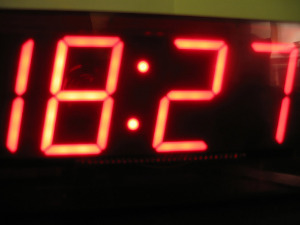} 
\caption{Redundant Clocks}
\label{fig:red_clocks2}
\end{subfigure}
~
\begin{subfigure}[b]{0.22\textwidth}
\centering
\fcolorbox{Green}{White}{\includegraphics[width=0.3\linewidth]{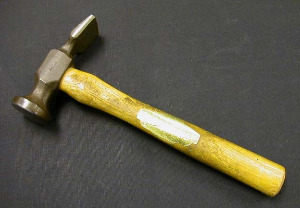}} 
\includegraphics[width=0.3\linewidth]{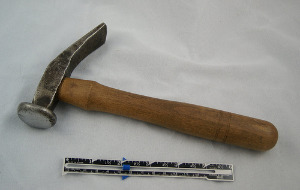} 
\includegraphics[width=0.3\linewidth]{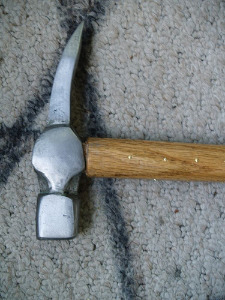} 
\caption{Redundant Hammers}
\label{fig:red_hammer2}
\end{subfigure}
~
\begin{subfigure}[b]{0.3\textwidth}
\centering
\fcolorbox{Green}{White}{\includegraphics[width=0.2\linewidth]{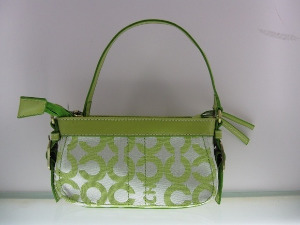}} 
\includegraphics[width=0.2\linewidth]{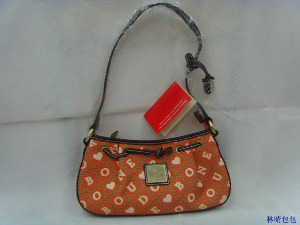} 
\includegraphics[width=0.2\linewidth]{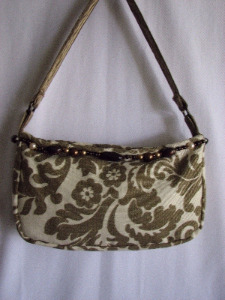} 
\includegraphics[width=0.2\linewidth]{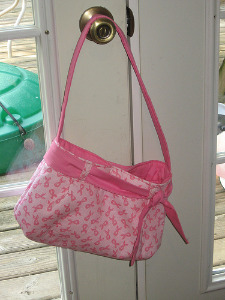}
\caption{Redundant Purses}
\label{fig:red_purses2}
\end{subfigure}
~
\begin{subfigure}[b]{0.22\textwidth}
\centering
\fcolorbox{Green}{White}{\includegraphics[width=0.3\linewidth]{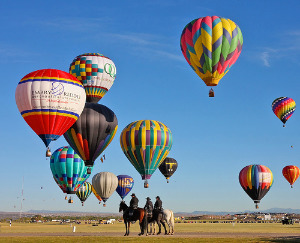}} 
\includegraphics[width=0.3\linewidth]{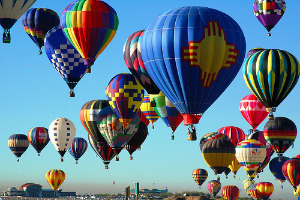}
\includegraphics[width=0.3\linewidth]{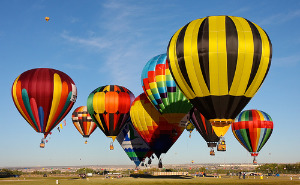}
\caption{Redundant Balloons}
\label{fig:red_balloon2}
\end{subfigure}
\vspace{0.1in}
\caption{
This figure highlights semantic similarities between images from the same redundant group and variation seen across different redundant groups of the same class. The redundant groups were found while creating
a 90\% subset of the \imagenet dataset.
Each sub-figure is a redundant group of images according to our algorithm.
Each column contains images belonging to the same class, with each row in a column
being a different redundant group.
For example, the first column contains the \emph{Clock}
class. Clocks in \ref{fig:red_clock1} are in one group of redundant images whereas clocks in \ref{fig:red_clocks2} are in another group.
From each of the groups in the sub-figures, only the images marked in green
boxes are selected by our algorithm and the others are discarded.
Discarding these images had no negative impact on validation accuracy.}
\label{fig:imagenet_var}
\end{figure*}

\begin{figure}[h]
\begin{subfigure}[b]{1\columnwidth}

\fcolorbox{Green}{White}{\includegraphics[width=0.3\linewidth]{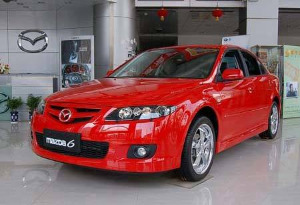}} 
\includegraphics[width=0.3\linewidth]{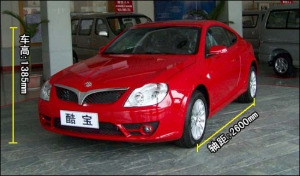} 
\hfill
\includegraphics[width=0.3\linewidth]{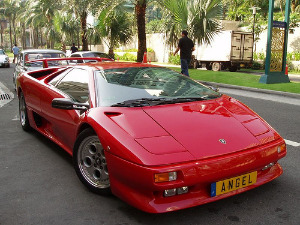} 
\label{fig:car_border}
\end{subfigure}

\begin{subfigure}[b]{1\columnwidth}
\fcolorbox{Green}{White}{\includegraphics[width=0.3\linewidth]{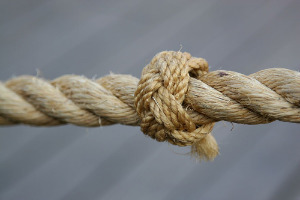}} 
\includegraphics[width=0.3\linewidth]{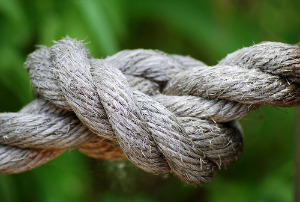} 
\hfill
\includegraphics[width=0.3\linewidth]{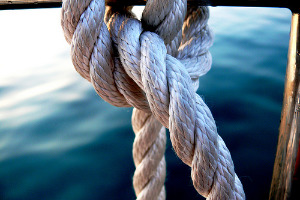} 
 \end{subfigure}

\begin{subfigure}[b]{1\columnwidth}
\fcolorbox{Green}{White}{\includegraphics[width=0.3\linewidth]{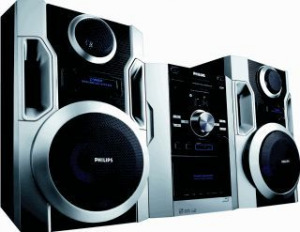}} 
\includegraphics[width=0.3\linewidth]{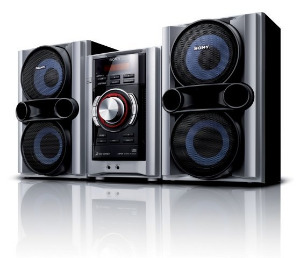} 
\hfill
\includegraphics[width=0.3\linewidth]{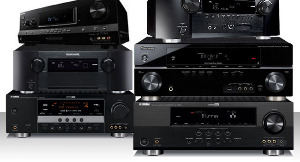} 

 \end{subfigure}

 \begin{subfigure}[b]{1\columnwidth}
\fcolorbox{Green}{White}{\includegraphics[width=0.3\linewidth]{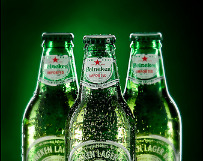}} 
\includegraphics[width=0.3\linewidth]{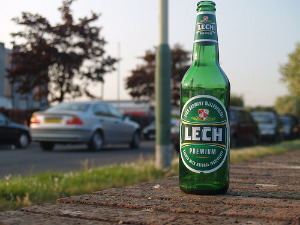} 
\hfill
\includegraphics[width=0.3\linewidth]{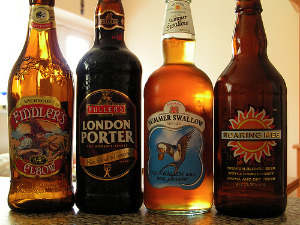} 

 \end{subfigure}

\caption{
In each row we plot two images from the same redundant group while creating a 90\%
subset on the left with the retained image highlighted in a green box.
On the right
we plot the image closest to the retained image in the semantic space but not included in the 
same redundant group.
Note that the image on the right shows a semantic variation which
is inconsistent with the one seen in the redundant group.}
\label{fig:imagenet_closest}
\end{figure}

The results of training with subsets of varying sizes of \imagenet dataset are shown in Figure \ref{fig:subset_imagenet}. Our proposed scheme is able to
successfully show that at least 10\% of the data can be removed from the training set without any negative impact on the validation accuracy,
whereas training on random subsets always gives a drop with decrease in subset size.

Figure \ref{fig:imagenet_dup} shows different redundant groups found in the \imagenet dataset.
It is noteworthy that the semantic change considered redundant is different across each group.
Figure \ref{fig:imagenet_var} highlights the similarities between images of the same redundant group and 
the variation across different redundant groups. 

In each row of Figure \ref{fig:imagenet_closest}, we plot
two images from a redundant group on the left where the retained image is
highlighted in a green box. On the right we display
the image closest to each retained image in dissimilarity but excluded from the redundant group.
These images were close in semantic space to the corresponding retained images, but were not considered similar enough to be redundant.
For example the redundant group in the first row of Figure \ref{fig:imagenet_closest} contains Sedan-like looking red cars. The 2-seater sports
car on the right, in spite of looking similar to the cars on the left, was not considered redundant with them.

Figure \ref{fig:sizes_imagenet} shows the number of redundant
groups of each size when creating a 90\% subset. Much akin to Figure
\ref{fig:sizes_cifar10}, a majority of images are not considered
redundant and form a group of size 1.

{\em Additional examples of redundancy group on ImageNet is provided in the appendix.}

\subsection{Implementation Details}
We use the open source Tensorflow \cite{tf} and tensor2tensor\cite{t2t} frameworks to train our models.
For clustering, we used the scikit-learn \cite{sklearn} library. For the \cifarten and \cifarhundred
experiments we train on a single NVIDIA Tesla P100 GPU.
For our \imagenet experiments we perform distributed
training on 16 Cloud TPUs.

\section{Conclusion}

In this work we present a method to find redundant subsets of training data.
We explicitly model a dissimilarity metric into our formulation which allows us 
to find semantically close samples that can be considered redundant.
We use an agglomerative clustering algorithm to find redundant groups
of images in the semantic space.
Through our experiments we are able to show that
\emph{at least} 
10\% of \imagenet and \cifarten datasets are redundant.

We analyze these redundant groups both qualitatively and quantitatively. 
Upon visual observation, we see that the semantic change considered redundant
varies from cluster to cluster. We show examples of a variety of varying
attributes in redundant groups, all of which are redundant from the
point of view of training the network.

One particular justification
for not needing this variation during training could be that
the network learns to be invariant to them because of its shared
parameters and seeing similar variations in other parts of the dataset.

In Figure  \ref{fig:subset_cifar10} and \ref{fig:subset_imagenet}, the accuracy without 5\% and 10\% of the
data is slightly higher than that obtained with the full dataset. This
could indicate that redundancies in training datasets hamper
the optimization process.

For the \cifarhundred dataset our proposed scheme fails to find any redundancies.
We qualitatively compare the redundant groups in \cifarhundred (Figure \ref{fig:cifar100_redundant}) to
the ones found in \cifarten (Figure \ref{fig:cifar_redundant}) and find that the semantic variation across
redundant groups is much larger in the former case. Quantitatively this
can be seen in Table \ref{tab:avg_dist} which shows points in redundant
groups of \cifarhundred are much more spread out in semantic space as compared
to \cifarten. 

Although we could not find any redundancies in the \cifarhundred dataset,
there could be 
a better algorithm that could find them. Moreover, we hope that this work
inspires a line of work into finding these redundancies and leveraging
them for faster  and more efficient training.

\section{Acknowledgement}
We would like to thank colleagues at Google Research for comments and discussions: Thomas Leung, Yair Movshovitz-Attias, Shraman Ray Chaudhuri, Azade Nazi, Serge Ioffe.

\clearpage

\bibliography{reduced_data}
\bibliographystyle{apalike}

\clearpage

\onecolumn
\appendix
\section{Appendix}
Each row is a redundant group of images. The left most image is retained in each row for the 90\% subset.

\begin{figure}[h!]
	\includegraphics[width=0.1\textwidth]{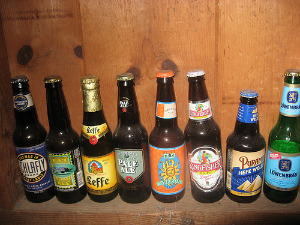}
	\includegraphics[width=0.1\textwidth]{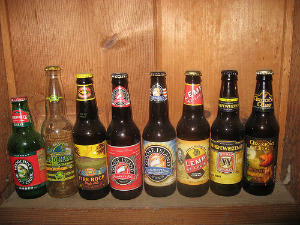}
	\includegraphics[width=0.1\textwidth]{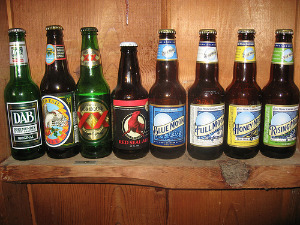}
	\includegraphics[width=0.1\textwidth]{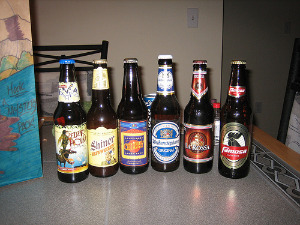}        \includegraphics[width=0.1\textwidth]{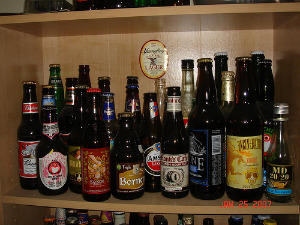}
	\includegraphics[width=0.1\textwidth]{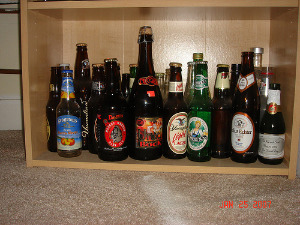}
	\includegraphics[width=0.1\textwidth]{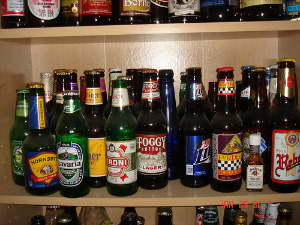}   
\end{figure}

\begin{figure}[h!]
	\includegraphics[width=0.1\textwidth]{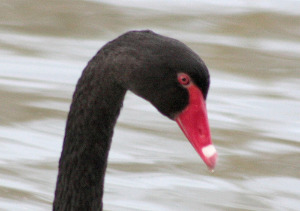}
	\includegraphics[width=0.1\textwidth]{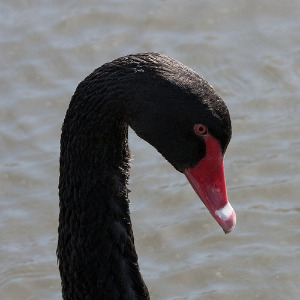}
	\includegraphics[width=0.1\textwidth]{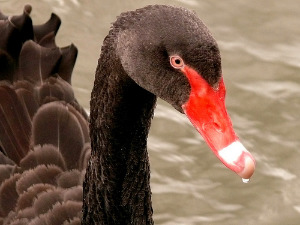}
	\includegraphics[width=0.1\textwidth]{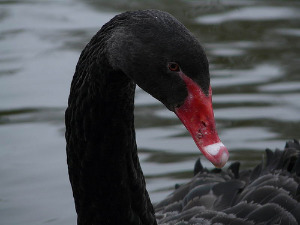}        \includegraphics[width=0.1\textwidth]{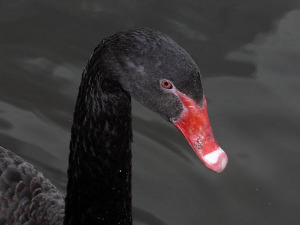}
\end{figure}

\begin{figure}[h!]
	\includegraphics[width=0.1\textwidth]{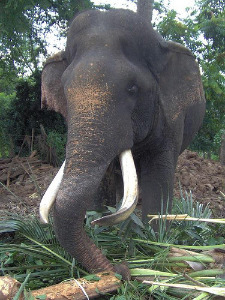}
	\includegraphics[width=0.1\textwidth]{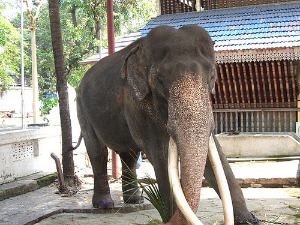}
	\includegraphics[width=0.1\textwidth]{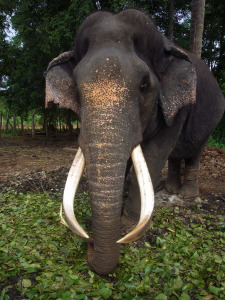}
	\includegraphics[width=0.1\textwidth]{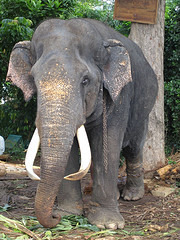}        \includegraphics[width=0.1\textwidth]{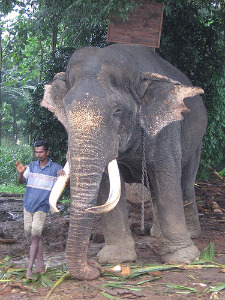}
\end{figure}

\begin{figure}[h!]
	\includegraphics[width=0.1\textwidth]{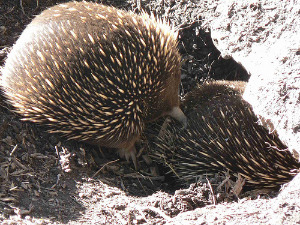}
	\includegraphics[width=0.1\textwidth]{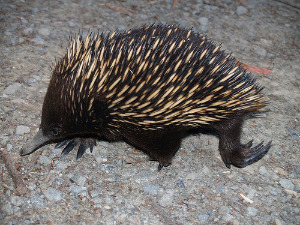}
	\includegraphics[width=0.1\textwidth]{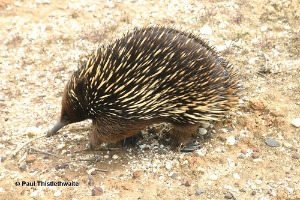}
	\includegraphics[width=0.1\textwidth]{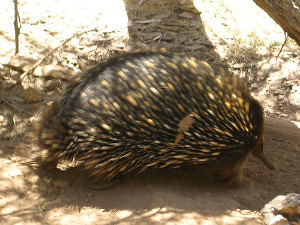}        
\end{figure}

\begin{figure}[h!]
	\includegraphics[width=0.1\textwidth]{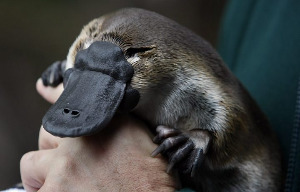}
	\includegraphics[width=0.1\textwidth]{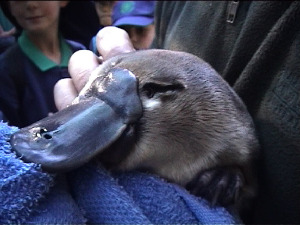}
	\includegraphics[width=0.1\textwidth]{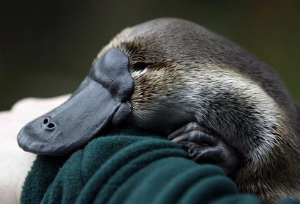}
\end{figure}

\begin{figure}[h!]
	\includegraphics[width=0.1\textwidth]{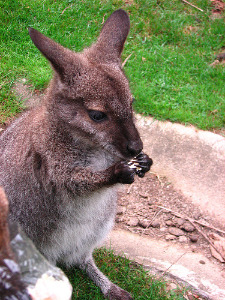}
	\includegraphics[width=0.1\textwidth]{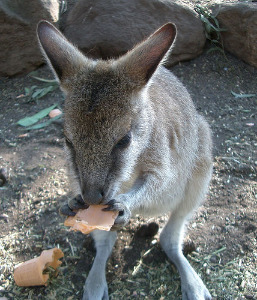}   
	\includegraphics[width=0.1\textwidth]{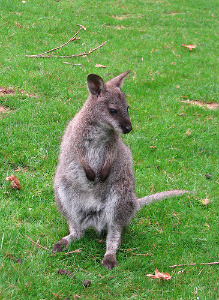}
	\includegraphics[width=0.1\textwidth]{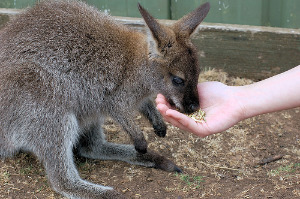}
	\includegraphics[width=0.1\textwidth]{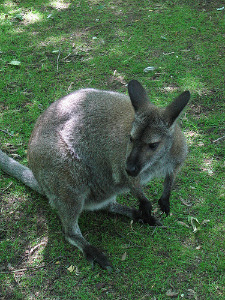}
	\includegraphics[width=0.1\textwidth]{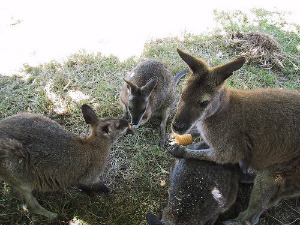}
	
\end{figure}
\begin{figure}[h!]
	\includegraphics[width=0.1\textwidth]{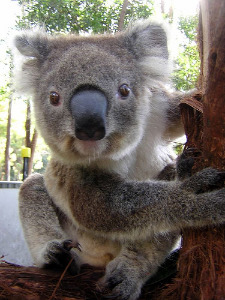}
	\includegraphics[width=0.1\textwidth]{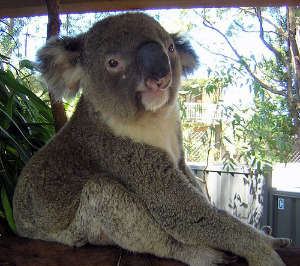}
	\includegraphics[width=0.1\textwidth]{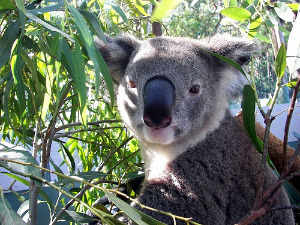}
	\includegraphics[width=0.1\textwidth]{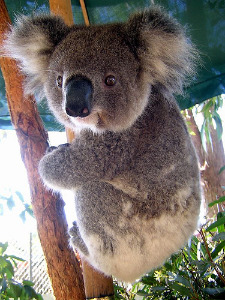}
\end{figure}

\begin{figure}[h!]
	\includegraphics[width=0.1\textwidth]{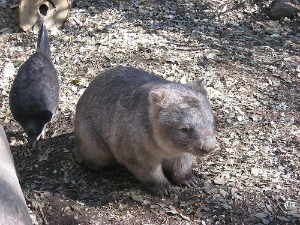}
	\includegraphics[width=0.1\textwidth]{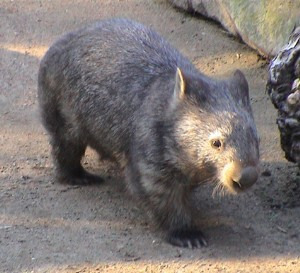}
	\includegraphics[width=0.1\textwidth]{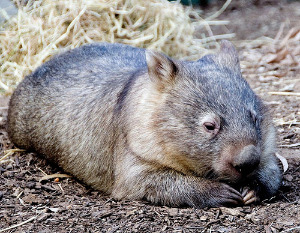}
	\includegraphics[width=0.1\textwidth]{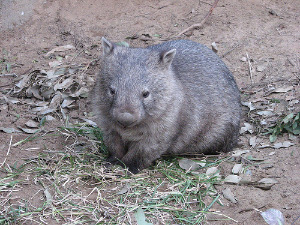}
	\includegraphics[width=0.1\textwidth]{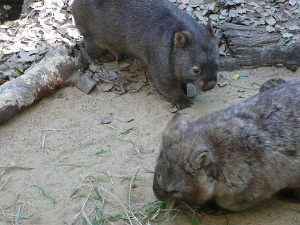}
	\includegraphics[width=0.1\textwidth]{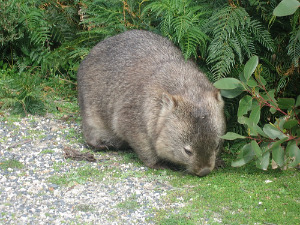}
\end{figure}

\begin{figure}[h!]
	\includegraphics[width=0.1\textwidth]{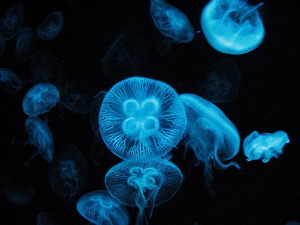}
	\includegraphics[width=0.1\textwidth]{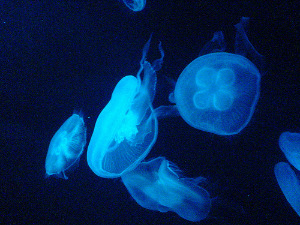}
	\includegraphics[width=0.1\textwidth]{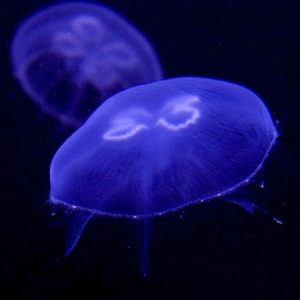}
	\includegraphics[width=0.1\textwidth]{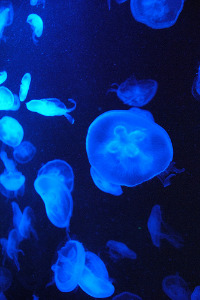}
	\includegraphics[width=0.1\textwidth]{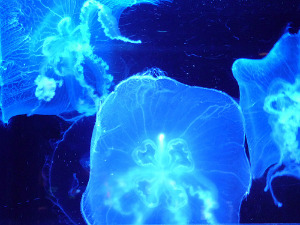}
	
\end{figure}

\begin{figure}[h!]
	\includegraphics[width=0.1\textwidth]{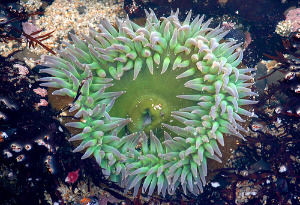}
	\includegraphics[width=0.1\textwidth]{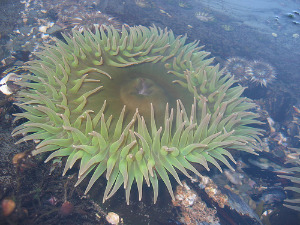}
	\includegraphics[width=0.1\textwidth]{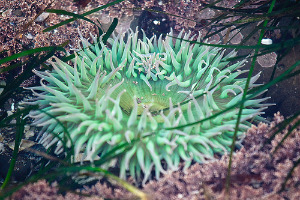}
	\includegraphics[width=0.1\textwidth]{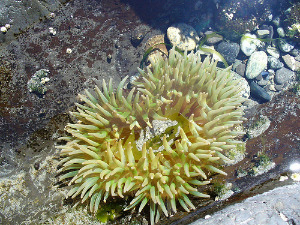}
\end{figure}

\begin{figure}[h!]
	\includegraphics[width=0.1\textwidth]{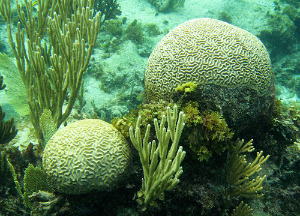}
	\includegraphics[width=0.1\textwidth]{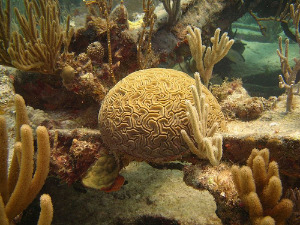}
	\includegraphics[width=0.1\textwidth]{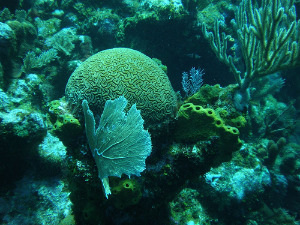}
	\includegraphics[width=0.1\textwidth]{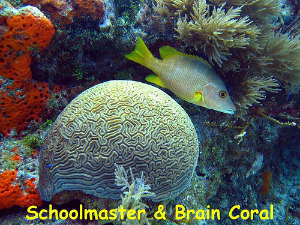}
\end{figure}
\begin{figure}[h!]
	\includegraphics[width=0.1\textwidth]{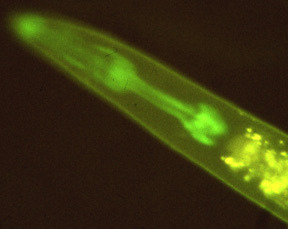}
	\includegraphics[width=0.1\textwidth]{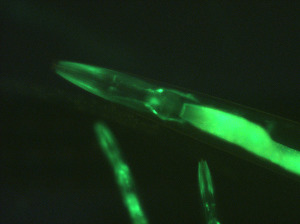}
	\includegraphics[width=0.1\textwidth]{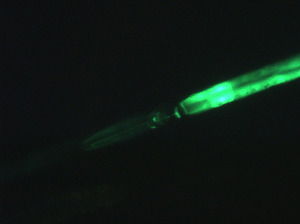}
	\includegraphics[width=0.1\textwidth]{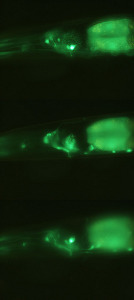}
	\includegraphics[width=0.1\textwidth]{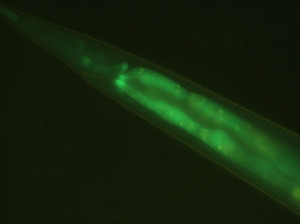}
\end{figure}
\vspace{100pt}
\

\end{document}